\pdfoutput=1

\documentclass[11pt]{article}

\usepackage[dvipsnames]{xcolor}

\usepackage{acl}

\usepackage{times}
\usepackage{latexsym}

\usepackage[T1]{fontenc}

\usepackage[utf8]{inputenc}

\usepackage{microtype}

\usepackage{amssymb}
\usepackage{algorithm}
\usepackage{algpseudocode}

\usepackage{placeins}

\usepackage{amsmath}
\usepackage{stfloats}
\usepackage{chngcntr}
\usepackage{makecell}
\usepackage{svg}
\usepackage{adjustbox}
\usepackage{subcaption}
\usepackage{multirow}
\usepackage{xurl}

\usepackage{enumitem}
\usepackage{multicol}
\usepackage{float}
\usepackage{circledsteps}
\usepackage[normalem]{ulem}

\newcommand{\gap}{\underline{\hspace{1cm}}}

\usepackage{pgfplots, pgfplotstable}
\pgfplotsset{compat=1.17}
\usepgfplotslibrary{external}
\usepackage[outline]{contour}
\usepackage[most]{tcolorbox}
\newtcbox{\unframedbox}[1]{
    on line,
    nobeforeafter,
    colback=#1,
    boxrule=0pt,arc=3pt,
    colframe=white,
    toprule=1pt,bottomrule=1pt,
    boxsep=0pt,left=1pt,right=1pt,top=0pt,bottom=0pt,
    equal height group=K,
    valign=center,
    before upper=\vrule width 0pt height 2ex depth 1ex\relax,
}
\newtcbox{\framedbox}[2]{
    on line,
    nobeforeafter,
    colback=#1,
    boxrule=1pt,arc=3pt,
    colframe=#2,
    boxsep=0pt,left=1pt,right=1pt,top=0pt,bottom=0pt,
    equal height group=K,
    valign=center,
    before upper=\vrule width 0pt height 2ex depth 1ex\relax,
}
\newcommand{\hl}[3][0]{\definecolor{bgcolour}{cmyk}{0,#2,#2,0}\setlength{\fboxsep}{0pt}\ifx0#1\unframedbox{bgcolour}{#3}\else\framedbox{bgcolour}{#1}{#3}\fi\hspace{-2pt}}%

\title{Constructing Open Cloze Tests Using Generation and Discrimination Capabilities of Transformers}

\author{
Mariano Felice, Shiva Taslimipoor and Paula Buttery \\
ALTA Institute, Computer Laboratory, University of Cambridge \\
Cambridge, UK \\
\texttt{\{mf501,st797,pjb48\}@cam.ac.uk} \\
} 

\begin{document}
\maketitle
\begin{abstract}

This paper presents the first multi-objective transformer model for constructing open cloze tests that exploits generation and discrimination capabilities to improve performance. Our model is further enhanced by tweaking its loss function and applying a post-processing re-ranking algorithm that improves overall test structure. Experiments using automatic and human evaluation show that our approach can achieve up to 82\% accuracy according to experts, outperforming previous work and baselines. We also release a collection of high-quality open cloze tests along with sample system output and human annotations that can serve as a future benchmark.

\end{abstract}

\defcitealias{ALTE:2005}{ALTE, 2005}
\defcitealias{ALTE:2011}{2011} 

\section{Introduction}

Open cloze \cite{taylor1953cloze} tests are a common type of exercise where words are removed from a piece of text and must then be filled in by the students without any options to choose from. They are often used in language learning environments as a quick and effective way to test vocabulary, grammar and reading comprehension \cite{tremblay2011proficiency,Trace:2020}. However, designing high-quality cloze tests for language learning is a laborious process that involves finding an optimal distribution of gaps based on aspects such as function, distance, number of answers, etc. \citepalias{ALTE:2005,ALTE:2011}.

In this paper, we propose a strategy to construct open cloze exercises using transformer models~\cite{vaswani2017}. Our transformer-based architecture employs two objectives to predict the words that should be gapped in a text passage.  
Our main  objective is standard token classification, where we aim to minimise the error of classifying a token as a gap or not. The second and auxiliary objective 
is a language-model-based objective whereby we attempt to minimise the language model error when predicting the right answer for each gap. Our solution is based on a pre-trained ELECTRA \citep{clark-etal-2020-electra} model that is fine-tuned on the two described objectives in a multi-task scenario.

Our output aims to mimic the style of open cloze tests in the First Certificate in English (FCE) exam\footnote{Now known as B2 First: \href{https://www.cambridgeenglish.org/exams-and-tests/first/}{\tt{https://www.cambridge\\english.org/exams-and-tests/first/}}}, which is targeted at learners of English at the B2 proficiency level of the Common European Framework of Reference (CEFR) for languages \cite{CEFR}. Unlike other tests, the FCE open cloze task aims to simultaneously test many aspects of grammar and vocabulary that students are expected to know at this level. Since the tests are created from a text passage, they must be skilfully designed in order to ensure an optimal distribution of gaps that adheres to guidelines. A shortened example is shown in \autoref{fig:fce-sample}. 

Our system is evaluated under two settings: 1) automatic evaluation, where the generated gaps are compared to gold-standard gaps proposed by test experts, and 2) human evaluation, where the quality of the generated gaps is judged by test experts.

\begin{figure*}[tb]
    \centering
    \small
    \begin{tabular}{|p{0.98\textwidth}|}
         \hline
         \\[-5pt] 
         \multicolumn{1}{|c|}{\textbf{Motorbike stunt rider}}\\[5pt]
         I work (1) \gap~ a motorbike stunt rider — that is, I do tricks on my motorbike at shows. The Le Mans race track in France was (2) \gap~ I first saw some guys doing motorbike stunts. I'd never seen anyone riding a motorbike using just the back wheel before and I was (3) \gap~ impressed I went straight home and taught (4) \gap~ to do the same. 
         \\[6ex]
         \hline
    \end{tabular}
    \caption{Sample FCE open cloze test (shortened).}
    \label{fig:fce-sample}
\end{figure*}

The main contributions of our work are as follows: 1) we are the first to employ transformer models for open cloze test generation, 2) unlike previous studies, we work at the paragraph level, which is a much more challenging task, 3) we propose a multi-task learning approach with two objectives: one is to classify tokens into gaps/non-gaps and the other to minimise the error of re-generating the gapped word, 4) we report state-of-the-art results, outperforming previous work and a strong baseline, 5) we propose additional components to control the structure of the final cloze tests as human experts do, 
6) we perform both automatic and human evaluation and 7) we make our test data, system output and human annotations available to the research community\footnote{Dataset available at \url{https://github.com/CambridgeALTA/fce-cep-oc}.}.

\section{Related Work}

While research into automatic cloze test generation is vast \citep{mostow2017,Kurdi-etal:2020,Yang2021}, work on open cloze tests for language learning is scarce. \citet{pino:2008} generate open cloze questions using sample sentences from a learners' dictionary based on four linguistic criteria: (grammatical) complexity, well-defined context (collocations), grammaticality and length. A later version of their system adds hints for gapped words \cite{pino:2009}.
Exercise Maker \citep{Malafeev:2014} is a rule-based open source system that attempts to emulate exercises in Cambridge English examinations based on the most frequently tested words. Most of the gaps it proposes were found to be useful and the automated exercises were hard to differentiate from authentic tests. 

\citet{Chinkina-etal:2017} generate open cloze exercises for phrasal verbs by extracting sentences from news articles and generating a pair of questions and answers where the identified particle verbs are gapped. Similarly, \citet{Soonklang-etal:2017} gap words in sentences according to their part of speech in order to practise articles, prepositions, etc. Finally, \citet{marrese-taylor-etal:2018} use LSTMs to build sequence labelling and classification models that decide where to insert a single gap in a single sentence. Automatic evaluation against gold-standard gaps showed the method was effective. 

Other work has focused on creating automated cloze tests by controlling aspects of the proposed gaps so that they correlate with a target proficiency level. \citet{Lee-et-al:2019}, for example, manipulate the difficulty of C-tests (open cloze tests with hints, \citet{c-test-2002}) by varying the position and word length of the gaps. A similar concept is presented by \citet{settles-etal-2020-machine} and \citet{mccarthy-etal-2021-jump}, although difficulty is predicted using a machine-learning model that correlates with CEFR levels. In these cases, tests are dynamically adapted to the examinee's proficiency level during the test session. From a different perspective, \citet{felice-buttery-2019-entropy} show that controlling gap entropy can be useful for designing open cloze tests at different CEFR levels. 
The work we present in this paper, however, aims to model the more complex task of predicting a full set of gaps at the paragraph level that comply with design and testing principles and is, 
to the best of our knowledge, the first to employ and adapt transformer-based models for this task.

System evaluation is also challenging, since there is usually more than one potential word in the text that could constitute a good gap. While previous work often made a choice between automatic \citep{marrese-taylor-etal:2018} or human evaluation \citep{Malafeev:2014,Das2017FactualOC} for their experiments, we perform both: automatic evaluation to identify the best models during development and human evaluation to measure test quality in the final output.

\section{Model}
\label{sec:model}

We define open cloze generation as the task of predicting a set of tokens that should be gapped in the text. Unlike previous approaches that work at the sentence level, our models work at the paragraph level (i.e. take the full text as input), since we believe the interactions between gaps can only be optimally captured when the text is processed as a whole rather than sentence by sentence.

\begin{figure}[t!]
	\resizebox{\columnwidth}{!}{
		\includegraphics{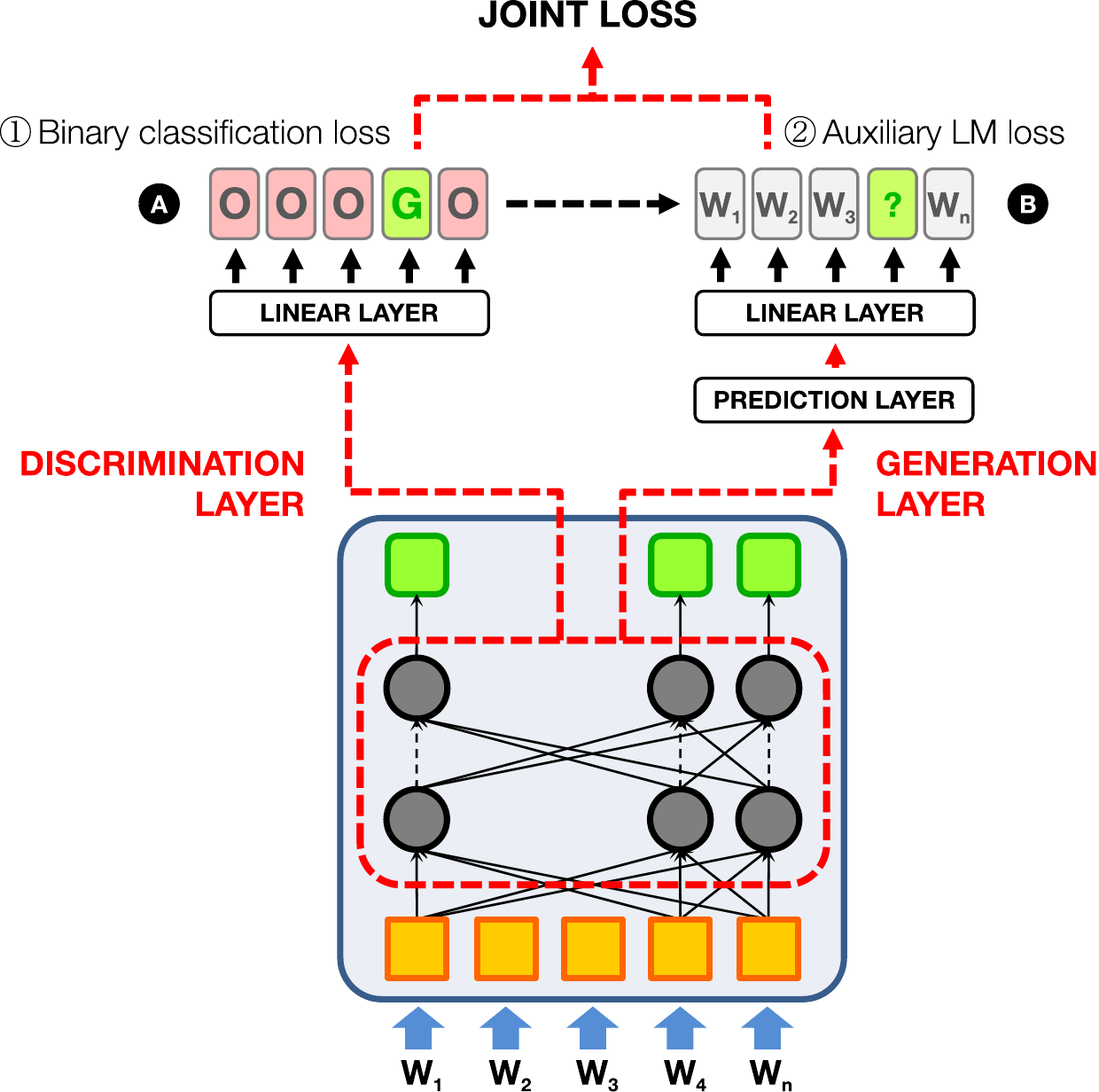}
	}
	\caption{Architecture of our multi-objective ELECTRA-based system. The model is simultaneously trained on two objectives: 1) token classification and 2) LM prediction of gapped words.}
	\label{fig:model}
\end{figure}

Given a text passage, we aim to predict the words that should be gapped in order to create a cloze test that would reliably assess student ability. The task is modelled as a supervised sequence tagging problem where each token is classified as being a good potential gap or not. We employ ELECTRA~\citep{clark-etal-2020-electra}, one of the state-of-the-art pre-trained transformer-based language representation models~\citep{wolf-etal-2020-transformers}. 
ELECTRA is an extension of BERT~\citep{devlin2018bert} with a different pre-training task which is a discriminator (rather than a generator) and aims to detect replaced tokens (rather than generating words for the masks).  
We believe that this discrimination objective makes it more suitable for our token classification task. Moreover, we also exploit ELECTRA's generation capabilities as a language model for estimating the answers to the proposed gaps as an auxiliary task. Hence, to make the most of this pre-trained model, we fine-tune it using two training objectives, as depicted in \autoref{fig:model}:

\begin{enumerate}[topsep=0.2ex,itemsep=0ex,partopsep=0.2ex,parsep=0.2ex,leftmargin=*,label=\protect\Circled{\arabic*}]
\item A \textit{token classification objective} which aims to minimise the error of classifying each token as a potential gap or not.
\item A \textit{language modelling objective} that aims to minimise the negative log-likelihood of re-generating the words that have been gapped.
\end{enumerate}

The first objective is typical of any standard token classification model and constitutes our key task. In particular, we use ELECTRA's discriminator head with softmax to tag each word in the input sequence as a `good' gap or not. All the gaps in our training data are replaced with the first intended target answer and labelled positive, while the remaining tokens are labelled negative (\Circled[fill color=black,inner color=white]{\textbf{A}}).

The second and auxiliary objective attempts to model our preference for gaps with a restricted number of answers while also ensuring that the original word can be guessed from the context. This is to avoid generating gaps that are too `open' and therefore ineffective, such as a gap that accepts any noun or adjective.
Specifically, we mask the words in the positions that are predicted as gaps by the discriminator and use ELECTRA's generative head to 
generate the expected words in the blanks (\Circled[fill color=black,inner color=white]{\textbf{B}}).
 
While the input layers
are shared between the discriminator and the generator model, the two branches of the system leading to the two objectives are fine-tuned in parallel 
in a multi-task setting. 

\section{Extensions}
\label{sec:extensions}

Our neural transformer-based sequence tagging model can be very effective at proposing potentially good gaps, but the task becomes more challenging when we expect the output to meet additional requirements such as no repetitions, no gap interdependence, a minimum distance between gaps and a varied selection of lexico-grammatical items. 
We address these issues using two complementary strategies: a manipulation of the loss function and a post-processing module.

\subsection{Loss manipulation}
\label{sec:loss}

In order to spread gaps evenly throughout the text, we modify the token-level loss function of our tagging model by imposing a higher penalty on tokens that are in close proximity to a gap. Let $g$ be the position of a gap in the sequence, then for each token in position $i$ in the proximity of $g$, i.e. $|g-i|<D$, the loss function ${l_i}'$ for the token in position $i$ is defined as:
\begin{equation}
{l_i}' = l_i * \frac{W}{|g-i|}
\label{eq:loss_manip}
\end{equation}

where $W$ represents the penalty and $D$ is the maximum distance scope for penalisation.\footnote{We empirically set the values of constants D and W to $3$ and $3.0$ respectively.} \autoref{eq:loss_manip} thus gives more weight to tokens closer to gaps, which results in higher penalisation of their cost functions whenever they are misclassified.

\subsection{Post-processing}
\label{sec:post}

We also employ a post-processing strategy where we replace the gaps that are repeated in the text with better options. We optimise the choice of these alternative gaps by considering the distance between them and the resulting distribution of gaps with different part-of-speech (PoS) tags. 

Our post-processing step can be seen as a re-ranking function. The gap candidates that are originally ranked based on the model's confidence scores change their ranking to match other desirable requirements of a well-structured cloze test.
If the selected \textit{n}-best gaps include repetitions, our post-processing algorithm randomly chooses one of them at a time and attempts to replace it with a better alternative. An alternative gap is deemed better if 1) its answer is not a repetition of another gapped word, 2) its distance to other selected gaps meets the minimum required distance or is higher than the pairwise distances of the originally selected gaps, and 3) it improves the PoS distribution of the gapped words. 
The PoS distribution of each new selection of gaps is compared to the average gapped PoS distribution of the cloze tests in the training data using Kullback-Leibler (KL) divergence. A combination of gaps that yields lower KL divergence is assumed to be a better solution.

\if{false}
\begin{algorithm*}
\small
\caption{Evaluating a candidate gap against an original gap chosen to be replaced}\label{alg:cap}
\begin{algorithmic}

\If{original\_gap == candidate\_gap:}
    \State \Return False
\EndIf
\If{candidate\_gap in selected\_gaps:}
  \State  \Return False
\EndIf
\State $min\_original\_distance \gets (compute\, the\, minimum\, pairwise\, distance\, between\, original\, gaps)$
\State $candidate\_min\_distance \gets (compute\, the\, minimum\, distance\, of\, the\, candidate\, to\, original\, gaps)$
\If{min\_original\_distance < 5 \& candidate\_min\_distance > min\_original\_distance}
\State  \Return True
\EndIf
\If{candidate\_min\_distance < 5}
  \State  \Return False
\EndIf
\If{KL\_divergance(original\_pos\_dist, reference\_pos\_dist) > KL\_divergance(candidate\_pos\_dist, reference\_pos\_dist)}
\State  \Return True
\EndIf
\end{algorithmic}
\end{algorithm*}
\fi

These extensions to the base model bring our final cloze tests closer to those created by human experts by automatically controlling variables that would otherwise need to be adjusted manually. This makes our solution a fully-automated system that can produce ready-to-use cloze tests from an input text passage.



\section{Data}
\label{sec:data}

To the best of our knowledge, there are no public datasets of full-text open cloze tests that could be used for our task. The CLOTH dataset \cite{CLOTH}, for example, contains gapped passages designed for language learners, but it is primarily focused on reasoning and reading comprehension and uses multiple choice questions where distractors play a major role, making it substantially different to the task we aim to model.

For this reason, we use a collection of expertly created open cloze tests at the B2 CEFR level that was kindly provided by Cambridge University Press \& Assessment (CUP\&A) for research purposes. Each task consists of a text passage of no more than $300$ tokens, a variable number of gaps (between $8$ and $16$) and a list of valid answers for each gap (between $1$ and $7$). During the design process, the tasks undergo extensive quality control and pretesting, so their gaps are guaranteed to be very effective at assessing student ability. 

For training, we reconstruct the texts by replacing each gap with its first answer and we split the whole collection into train, dev and test. Details of our dataset are shown in  \autoref{tab:datasets}.

Given the lack of publicly available data, we make our test set available with this paper so as to provide a common benchmark for the task and to encourage further research in this area. All the texts were tokenised and parsed using spaCy v2.3\footnote{\url{https://spacy.io/}}.

\begin{table}[t]
    \centering
    \begin{tabular}{l|rrr}
                    & \textbf{Train} & \textbf{Dev} & \textbf{Test}  \\ \hline
         Tasks      &  356  & 58 & 36 \\
         Tokens   &  79,863  & 12,797  & 6,621 \\
         Gaps & 4,565 & 787  & 360 \\
    \end{tabular}
    \caption{\label{tab:datasets} Number of tasks, tokens and gaps in each section of the data.}
\end{table}

\section{Experiments}
\label{sec:experiments}

\subsection{Setup}
We use the pre-trained ELECTRA base discriminator model\footnote{\url{https://github.com/huggingface/transformers}.} with $12$ attention heads and $12$ hidden layers. Along with all the tokens in the sequences, we also input dependency parsing information to the system. More specifically, we concatenate the ELECTRA representation of each token with the representation of its head in the dependency graph.\footnote{If the token is the head, then its representation is repeated.} On top of the encoding layers, we have two branches that are being learned simultaneously (\autoref{fig:model}). 

The first branch is a simple linear layer that aims to classify each token as a gap or non-gap. For the second branch, we add ELECTRA's generation layer plus a linear layer which aims to predict the best word from the whole vocabulary as an auxiliary task. We are only interested in predicting the answer words for the gaps. Therefore, we change the input to the second branch by masking the words that are predicted as gaps by the first branch at each step of training.
We employ cross-entropy loss on each branch and ignore the loss values for the tokens that are not masked in the second branch. The whole architecture is updated based on the sum of the two losses. Fine-tuning parameters are specified in \autoref{app:paramaters}.

\subsection{Baselines}

We compare our multi-objective ELECTRA model 
to other systems, namely:
\begin{description}[topsep=0.5ex,itemsep=0ex,partopsep=0.5ex,parsep=0.5ex]
\item[Random baseline] Generates a random set of gaps for each task 
based on the average probability distribution of gapped PoS in the training data.
\item[Exercise Maker] Generates gaps using rules and a pre-compiled list of commonly gapped words from a variety of Cambridge English main suite exams \citep{Malafeev:2014}. Set to FCE mode for our experiments.
\item[BERT] Predicts potentially good gaps using BERT \cite{devlin2018bert} for token classification. We use the pre-trained base model with standard parameters and fine-tune the weights of the whole architecture. 

\item[Standard ELECTRA] Similar to BERT, it predicts potentially good gaps using a standard pre-trained ELECTRA-base model. 
This is a single-objective model that is fine-tuned on token classification only. 

\end{description}

Both \textbf{random} and \textbf{Exercise Maker} attempt to generate the same number of gaps per task as defined in the gold standard, although this is not always possible since the required conditions  (such as specific words or PoS) are not always met.

\subsection{Evaluation}

\begin{table*}[ht!]
\centering
\small
\begin{tabular}{llp{10cm}}
\hline
\textbf{Class} & \textbf{Label} & \textbf{Description} \\
\hline
Good & Good & The gap is appropriate, i.e. it is expected to be effective during testing. \\
Bad & Too close to other gaps & The gap is in close proximity to another gap. \\
Bad & Too many possible answers & The gap allows too many answers (often more than $5$). \\
Bad & Too many gaps of this type & There are many gaps with the same part-of-speech or testing focus in the text. \\
Bad & Answers can change meaning & The gap can be filled by answers that would change the meaning of the text, e.g. `and' or `but'. \\
Bad & Answers can have different PoS & The gap can be filled by answers that have a different grammatical function, e.g. `which' or `and'. \\
Bad & Gap depends on another & There is some dependency between this gap and another in the text. \\
Bad & Repeated gap & There is already another gap testing the same word in the text. \\
Bad & Phantom gap & The gap does not require an answer for the text to make sense. \\
Bad & Unacceptable outlier & The gap does not fit in the text for multiple reasons (e.g. inappropriate difficulty). \\
Bad & Other (please specify) & Any other reason why the gap is considered unsuitable. \\
\end{tabular}
\caption{\label{tab:gap-labels}
Labels used in human annotation.
}
\end{table*}

We report precision (P), recall (R) and F\textsubscript{1} scores based on a strict matching between the gaps predicted by our models and those in the gold standard. While this evaluation strategy might seem 
strict, it has the advantage of being fully automatic, thus avoiding the subjectivity and time required by human evaluation, so we adopt it during development.

In addition to letting the models decide the optimal number of gaps, we also evaluate system performance when we fix
the number of predicted gaps for each task to the number of gaps they have in the gold standard. The \textit{n}-best predicted gaps are chosen based on their confidence scores. In this scenario, P, R and F\textsubscript{1} become the same. 



We also report human evaluation by three test experts from CUP\&A who volunteered for the task. The experts were asked to label each proposed gap in each task of our test set (a total of $360$ gaps) as either good or bad and provide a reason and optional comments for their choice. The list of labels available to our annotators is shown in \autoref{tab:gap-labels}.

\section{Results and Discussion}

\subsection{Automatic evaluation}

\begin{table}
\centering
\begin{adjustbox}{width=1\columnwidth}
\begin{tabular}{llllllllc}
\hline
\textbf{Model} & \multicolumn{1}{c}{\textbf{P}} & \multicolumn{1}{c}{\textbf{R}} & \multicolumn{1}{c}{\textbf{F\textsubscript{1}}} \\
\hline
Random baseline & 15.29 & 14.87 & 15.08 \\
Exercise Maker & 23.33 & 25.79 & 24.50 \\
BERT & 51.16 & 47.65 &	49.34 \\
Standard ELECTRA & 55.61 & 46.00 & 50.35 \\
Multi-objective ELECTRA &  57.41 & 46.25 &	\textbf{51.23}
\\
\hline
\end{tabular}
\end{adjustbox}
\caption{\label{tab:dev-results}
Models' performance on the development set. 
}
\end{table}

We carry out automatic evaluation by computing P, R and F\textsubscript{1} on our development set. \autoref{tab:dev-results} reports the results of our multi-objective ELECTRA model (enhanced with dependency information) as well as the random baseline, Exercise Maker, BERT, and the standard single-task ELECTRA. This is our base model, which does not include any loss manipulation or post-processing. In this setting, the number of predicted gaps was decided by each model based on the confidence scores (${>0.5}$ for the positive class).

\begin{figure}[t!]
\centering
	\resizebox{6cm}{!}{
		\includegraphics{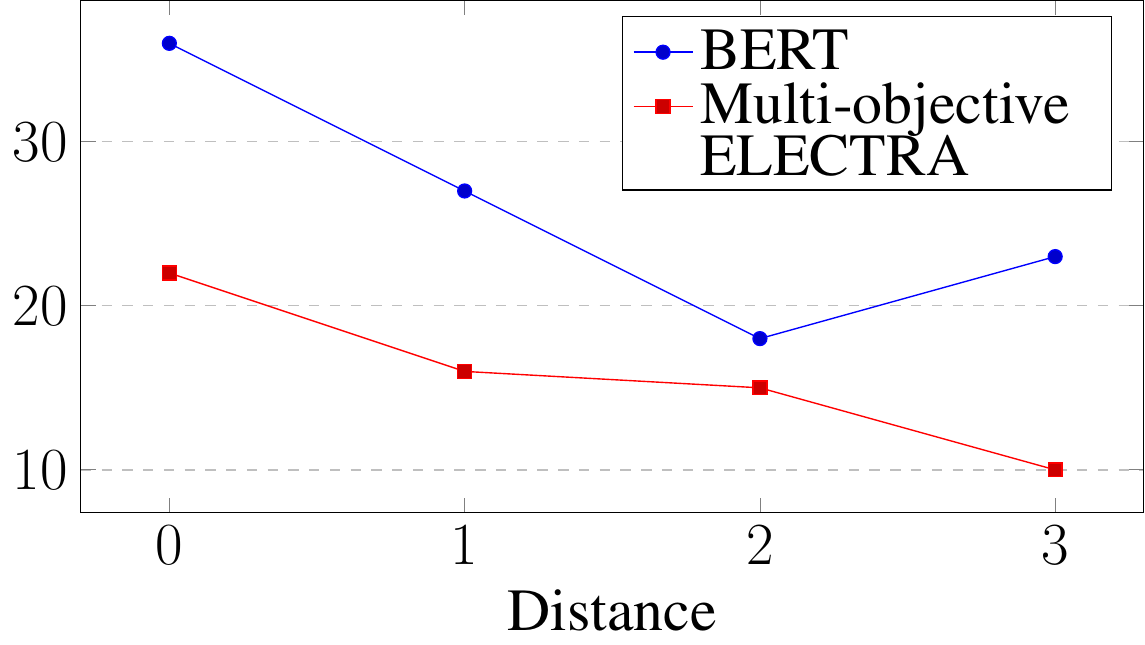}
	}
	\caption{Frequency of pairs of gaps with distance ranging from $0$ to $3$. Distance is measured by the number of words in between two gapped words. The minimum acceptable number of words between two gaps is $4$.}
	\label{fig:dist-chart}
\end{figure}

Overall, we observe that performance increases with more sophisticated models. Exercise Maker relies on previously seen gaps and so outperforms the random baseline by a large margin. However, it can only create gaps for the $139$ words in its predefined FCE word list, missing gaps that are not on that list.
Neural transformer-based models are the best, with improvements over Exercise Maker of at least $25$ F\textsubscript{1} on our development set. Although the improvement of our multi-objective ELECTRA model over BERT does not seem to be very significant based simply on P, R and F\textsubscript{1}, a closer look at the results reveals that BERT produces a much higher number of repeated gaps ($25$ compared to $9$ by multi-objective ELECTRA) as well as more cases of gaps in close proximity, as shown in \autoref{fig:dist-chart}.\looseness=-1

We also perform an ablation study in \autoref{tab:dev-results} where we compare our multi-objective ELECTRA model to a standard one that does not include our auxiliary language model objective. Results show that the former outperforms the latter on all metrics, confirming that the addition of the LM objective is clearly beneficial.

\autoref{tab:p_at_n} shows the performance of our multi-objective ELECTRA model as we increase the \textit{n}-best list of gaps according to their confidence score. The first row indicates the results of the system when it is forced to predict the exact same number of gaps per task as in the gold standard.\footnote{The number of gaps can vary per passage (see \autoref{app:dataset}).}
This causes P and R to be the same. As we expect, the results show that the number of gaps in the gold data is actually the optimal number to achieve the best F\textsubscript{1} score.


Although our multi-objective model shows good performance based on automatic evaluation, a closer look at the output reveals that the structure of the cloze tests is far from ideal as they often contain repetitions and gaps that are too close to each other, aspects that are carefully controlled in the gold standard. \autoref{tab:loss-manip} shows that system performance effectively improves as we add the 
extensions proposed in \autoref{sec:extensions}, indicating that global aspects of the task are not properly captured by our initial model and require further manipulation.


\begin{table}
\centering
\begin{tabular}{cccc}
\hline
\textbf{\# of predicted gaps} & \textbf{P} & \textbf{R} & \textbf{F\textsubscript{1}} \\
\hline
As-in-gold & 54.26 &	54.26 &	\textbf{54.26} \\
10 & 56.72 &	42.80 &	48.14 \\
15 & 51.49 &	56.93 &	54.07 \\
20 & 44.83 &	66.07 &	53.42 \\
30 & 35.63 & 78.78 & 49.07 \\
\end{tabular}
\caption{\label{tab:p_at_n}
Results of multi-objective ELECTRA when we predefine the number of predicted gaps.
}
\end{table}

\begin{table}
\centering
\begin{adjustbox}{width=1\columnwidth}
\begin{tabular}{p{4.5cm}ccc}
\hline
~ & \textbf{P} & \textbf{R} & \textbf{F\textsubscript{1}} \\
\hline
Multi-objective ELECTRA & 57.41 & 46.25 & 51.23 \\
+ loss manipulation & 47.87 &	59.85 &	\textbf{53.19} \\
+ post-processing & 48.42 & 60.23 & \textbf{53.68} \\
\hline
\end{tabular}
\end{adjustbox}
\caption{\label{tab:loss-manip}
Effect of loss manipulation  and post-processing on our multi-objective ELECTRA model. 
}
\end{table}


In order to make the structure of our output as similar as possible to our target tasks,
we fix the number of predicted gaps for each task to the number of gaps they have in the gold standard.  
Note that P and R are the same in this setting so we only report F\textsubscript{1}. 
The effect of this decision is shown is \autoref{tab:postprocessing}.
We can see that adding loss manipulation to our model decreases the number of adjacent gaps from $40$ to $23$, but increases the number of repeated gapped words from $18$ to $33$.  The decline in the restricted F\textsubscript{1} based on automatic evaluation is not favourable, but we make this sacrifice at the price of achieving a better-structured final test. 

\begin{table}[t]
\centering
\begin{adjustbox}{width=1\columnwidth}
\setlength{\tabcolsep}{1pt}
\begin{tabular}{|p{4.2cm}|c|c|c|}
\hline
~ & \textbf{Restricted} & \textbf{Repeated} & \textbf{Adjacent} \\ 
 & \textbf{F\textsubscript{1}} & \textbf{gaps} & \textbf{gaps}  \\ 
\hline
Multi-objective ELECTRA & 54.26 & 18 & 40 \\
+ loss manipulation & 51.59 & 33 & 23  \\
+ post-processing & 51.33 & \textbf{9} & \textbf{23}\\
\hline
\end{tabular}
\end{adjustbox}
\caption{\label{tab:postprocessing} Analysis of our model after adding extensions: loss manipulation and post-processing.
}
\end{table}


After adding post-processing for repeated gaps, we observe that, although overall F\textsubscript{1} performance drops slightly, the number of repeated gapped words decreases favourably from $33$ to $9$ (\autoref{tab:postprocessing}).
It also creates a better spread of gaps, as shown by a lower KL-divergence between the average PoS distribution of the output and that of the gold standard ($0.55$ with post-processing as opposed to $0.59$ without it). Post-processing also removes two cases in the development set where the gaps do not meet the minimum 4-word distance. 

It is worth recalling that these extensions are highly effective when we do not restrict the number of predicted gaps. \autoref{tab:loss-manip} shows that they significantly improve R, which results in higher overall F\textsubscript{1}. 

As a result of these experiments, we stick with our post-processing approach for the rest of our experiments and use it to produce the output submitted for human annotation.




\subsection{Human evaluation}
\label{sec:human-eval}
Following our intuition that test experts could find more value in our system than initially shown by our automatic evaluation, we asked a panel of three test experts to judge the quality of the gaps produced by our extended model on the test set. 
Inter-annotator agreement on gap classification (good/bad) was found to be moderate (percent agreement is $75.93\%$, Randolph's free-marginal kappa is $0.52$ \citep{Randolph2005}).



Unlike in automatic evaluation, we only report accuracy for our human experiment. System performance using automatic and human evaluation is compared in \autoref{tab:auto-vs-human} (reported individually for each annotator). These results show that performance increases dramatically when the output is judged by human experts, confirming our suspicion that performance is underestimated by automatic evaluation and that there are many other words in the texts that could constitute equally useful gaps apart from those in the gold standard. With system accuracy ranging between $75\%-82\%$ for human judgements, we can conclude that at least $7$ out of $10$ gaps proposed by our system are considered good by our experts.

\begin{table}
\centering
\begin{tabular}{cccc}
\hline
\textbf{Auto} & \textbf{Ann. 1} & \textbf{Ann. 2} & \textbf{Ann. 3} \\
\hline
53.89 & 82.50 & 75.83 & 77.50 \\
\hline
\end{tabular}
\caption{\label{tab:auto-vs-human}
Accuracy of our extended multi-objective ELECTRA model based on automatic and human evaluation on the test set.
}
\end{table}

We observed that differences between annotators' judgements and the gold-standard can occur for many reasons, e.g.: 
\begin{itemize}[topsep=0ex,itemsep=0ex,partopsep=0ex,parsep=0ex]
\item non-gaps in the gold standard are not necessarily bad gaps,
\item gold standard gaps are derived from pilot testing while annotators' gaps are derived from their expertise,
\item previous judgements by the annotators can affect the judgement of new gaps (e.g. choosing the best of two close gaps), etc. 
\end{itemize}
Annotator accuracy against the gold standard ranges between $50\%-60\%$.

\begin{figure}[t!]
	\resizebox{\columnwidth}{!}{
	    \includegraphics{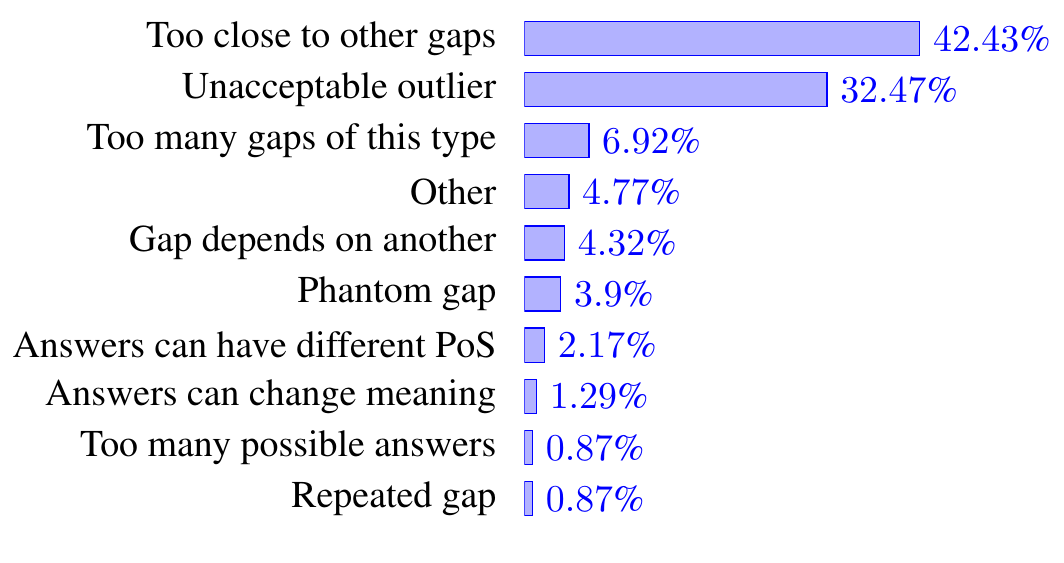}
	}
	\caption{Average frequency of the reasons given by the annotators for rejecting a gap.}
	\label{fig:bad-reasons}
\end{figure}

Following our classification in \autoref{tab:gap-labels}, we analysed the reasons why some gaps were not considered good by the annotators. \autoref{fig:bad-reasons} shows the average frequency of the different reasons given by the annotators for rejecting a gap proposed by our system. Examples are included in \autoref{app:reasons}. 

The most frequent reason is the violation of the minimum required distance between two gaps ($42.43\%$). Although our loss-manipulation approach was successful in reducing these cases, we did not attempt to eradicate them completely since there are many factors at play when choosing more appropriate gaps than just distance. In many cases, gaps in close proximity test different words in the same phrase (e.g. \textit{\uline{take} \uline{part} in}, \textit{\uline{in} addition \uline{to}}, etc.) so we preferred to keep these cases and encourage annotators to comment on their preferences.
Repetitions, on the contrary, are much better handled, accounting for only $0.87\%$ of all bad gaps.

The second most frequent reason is `unacceptable outlier' ($32.47\%$), which normally accounts for cases where the difficulty of the gap is considered inappropriate for the target proficiency level (B2 in this case). This is an interesting phenomenon, since the fact that the text as a whole pertains to a given CEFR level does not guarantee that the gaps created will always be appropriate for the level. The remaining reasons are substantially less frequent than the first two and mostly related to aspects that were not explicitly controlled in our models, except for the third topmost reason (`Too many gaps of this type') which we did control by comparing PoS distributions. These results show that our system is able to capture many aspects of the task that were not explicitly modelled.

Finally, we compared system accuracy per task computed from annotators' judgements vs. the gold standard. Average correlation across all annotators was found to be very weak (Pearson's $r=0.0558$, Spearman's $\rho=0.1474$).
This suggests that automatic scores are not a good proxy for human perception, with experts being much more positive about our model's output (as shown in \autoref{tab:auto-vs-human}).


\begin{figure*}[t]
    \centering
    \small
    \begin{tabular}{|p{0.98\textwidth}|}
         \hline
         \\[-5pt] 
        \multicolumn{1}{|c|}{\textbf{Gardening}}\\[5pt]
        \hl{0.247}{It} \hl{0.190}{is} \hl{0.074}{early} \hl{0.017}{summer} \hl{0.139}{,} \hl{0.239}{the} \hl{0.067}{season} \hl{0.232}{of} \hl{0.035}{abundance} \hl{0.081}{,} \hl{0.822}{when} \hl{0.123}{my} \hl{0.002}{garden} \hl{0.193}{is} \hl[black]{0.858}{\color{yellow}{\contour{red}{at}}} \hl{0.810}{its} \hl{0.050}{fullest} \hl{0.045}{.} \hl{0.003}{Flowers} \hl{0.202}{are} \hl[black]{0.834}{\color{yellow}{\contour{red}{in}}} \hl{0.015}{bloom} \hl{0.145}{and} \hl{0.069}{the} \hl{0.000}{grass} \hl{0.151}{is} \hl{0.030}{growing} \hl[black]{0.954}{\color{yellow}{\contour{red}{so}}} \hl{0.114}{fast} \hl{0.128}{that} \hl{0.131}{half} \hl{0.182}{an} \hl{0.079}{hour} \hl[black]{0.882}{after} \hl{0.005}{cutting} \hl{0.178}{it} \hl{0.018}{,} \hl{0.103}{I} \hl{0.166}{seem} \hl{0.208}{to} \hl{0.244}{be} \hl{0.136}{back} \hl[black]{0.942}{\color{yellow}{\contour{red}{where}}} \hl{0.109}{I} \hl{0.076}{started} \hl{0.042}{.} \hl{0.213}{This} \hl{0.108}{year} \hl[black]{0.894}{for} \hl{0.714}{the} \hl{0.192}{first} \hl{0.173}{time} \hl{0.133}{I} \hl{0.205}{am} \hl{0.059}{attempting} \hl{0.176}{to} \hl{0.020}{grow} \hl[black]{0.870}{\color{yellow}{\contour{red}{my}}} \hl{0.155}{own} \hl{0.010}{vegetables} \hl{0.092}{,} \hl{0.240}{an} \hl{0.082}{attempt} \hl{0.750}{that} \hl{0.217}{has} \hl{0.690}{so} \hl{0.666}{\color{yellow}{\contour{red}{far}}} \hl{0.163}{proved} \hl{0.124}{very} \hl{0.126}{successful} \hl{0.064}{.} \hl{0.096}{My} \hl{0.007}{vegetable} \hl{0.034}{plants} \hl{0.230}{have} \hl{0.227}{been} \hl{0.089}{yielding} \hl{0.198}{an} \hl{0.049}{abundance} \hl{0.165}{of} \hl{0.040}{produce} \hl{0.094}{,} \hl{0.738}{in} \hl{0.225}{fact} \hl{0.141}{much} \hl[black]{0.918}{\color{yellow}{\contour{red}{more}}} \hl{0.774}{than} \hl{0.097}{I} \hl{0.197}{can} \hl{0.134}{possibly} \hl{0.029}{consume} \hl{0.158}{myself} \hl{0.054}{.} \hl{0.111}{I} \hl{0.116}{’m} \hl{0.071}{convinced} \hl{0.129}{that} \hl{0.106}{you} \hl{0.702}{cannot} \hl{0.039}{plant} \hl{0.203}{even} \hl{0.220}{a} \hl{0.156}{single} \hl{0.008}{tomato} \hl[black]{0.930}{\color{yellow}{\contour{red}{without}}} \hl{0.022}{feeling} \hl{0.150}{a} \hl{0.055}{connection} \hl{0.148}{to} \hl{0.121}{the} \hl{0.044}{earth} \hl{0.187}{and} \hl{0.146}{to} \hl{0.188}{the} \hl{0.086}{countless} \hl{0.113}{generations} \hl{0.798}{who} \hl{0.229}{have} \hl{0.077}{worked} \hl{0.171}{the} \hl{0.072}{land} \hl{0.245}{before} \hl{0.175}{you} \hl{0.091}{.} \hl{0.195}{To} \hl{0.047}{plant} \hl{0.027}{seeds} \hl{0.678}{and} \hl{0.170}{then} \hl{0.161}{to} \hl{0.025}{harvest} \hl{0.786}{what} \hl{0.084}{you} \hl{0.212}{have} \hl{0.052}{grown} \hl{0.183}{gives} \hl{0.237}{\color{yellow}{\contour{red}{a}}} \hl{0.057}{deep} \hl{0.101}{sense} \hl{0.218}{of} \hl{0.087}{satisfaction} \hl{0.061}{.} \hl{0.143}{I} \hl{0.099}{believe} \hl{0.160}{that} \hl{0.153}{many} \hl{0.013}{doctors} \hl{0.185}{and} \hl{0.024}{mental} \hl{0.062}{health} \hl{0.066}{organisations} \hl{0.242}{all} \hl[black]{0.906}{\color{yellow}{\contour{red}{around}}} \hl{0.207}{the} \hl{0.032}{world} \hl{0.118}{now} \hl{0.138}{recognise} \hl{0.224}{the} \hl{0.119}{value} \hl{0.235}{of} \hl{0.012}{gardening} \hl{0.222}{to} \hl{0.200}{the} \hl{0.180}{well-being} \hl{0.215}{of} \hl{0.210}{those} \hl{0.234}{\color{yellow}{\contour{red}{who}}} \hl[black]{0.846}{take} \hl{0.762}{part} \hl{0.726}{in} \hl{0.168}{this} \hl{0.037}{activity} \hl{0.104}{.} 
        \\[10ex]
         \hline
    \end{tabular}
    \caption{Sample output of our extended ELECTRA model. Darker shades of red indicate higher confidence in inserting a gap. Predicted gaps are framed in black while gold standard gaps are in yellow font.}
    \label{fig:heatmap}
\end{figure*}

\subsection{Predictions by Gapped Word Frequency}


We found that our model does not overfit to words that are most frequently gapped in the training data, with correlation between gapped word frequency and F\textsubscript{1} scores in the test set being negligible (Pearson's $r=0.0108$, Spearman's $\rho=0.0915$).

Interestingly, while our model was unable to predict gaps not previously seen in the training data (\textit{turned}, \textit{amount}, \textit{pushed} and \textit{started}), it did predict a (previously unseen) gap for the word \textit{fewer}, which did not match the gold standard but was unanimously deemed good by our annotators.

\subsection{Predictions by PoS}

\begin{table}[t!]
\centering
\small
\begin{adjustbox}{width=1\columnwidth}
	\begin{tabular}{|l|r|rrr|}
		\hline
        \multirow{2}{*}{\textbf{PoS}} & 
        \multicolumn{1}{c|}{\textbf{Proportion}} & 
        \multicolumn{1}{c}{\multirow{2}{*}{\textbf{P}}} & 
        \multicolumn{1}{c}{\multirow{2}{*}{\textbf{R}}} & 
        \multicolumn{1}{c|}{\multirow{2}{*}{\textbf{F\textsubscript{1}}}} \\
        ~ & \textbf{in TEST} & ~ & ~ & ~ \\
		\hline
        ADP	&	20.59\%	&	50.00	&	43.24	&	46.38	\\
        ADV	&	14.17\%	&	57.69	&	58.82	&	58.25	\\
        DET	&	13.89\%	&	56.41	&	44.00	&	49.44	\\
        SCONJ	&	13.89\%	&	59.09	&	78.00	&	67.24	\\
        AUX	&	10.83\%	&	45.83	&	28.21	&	34.92	\\
        PRON	&	9.44\%	&	47.92	&	67.65	&	56.10	\\
        ADJ	&	4.44\%	&	60.00	&	75.00	&	66.67	\\
        NOUN	&	3.33\%	&	77.78	&	58.33	&	66.67	\\        
        NUM	&	2.78\%	&	61.54	&	80.00	&	69.57	\\
        CCONJ	&	2.50\%	&	55.56	&	55.56	&	55.56	\\
        VERB	&	2.22\%	&	50.00	&	50.00	&	50.00	\\
        PART	&	1.67\%	&	0.00	&	0.00	&	0.00	\\
        INTJ	&	0.28\%	&	50.00	&	100.00	&	66.67	\\
        \hline
	\end{tabular}
	\end{adjustbox}
	\caption{\label{tab:pos-results} Performance by PoS on the test set based on automatic evaluation.}
\end{table}

We also classified predictions based on their PoS tags\footnote{Using the Universal Dependencies tagset: \url{https://universaldependencies.org/u/pos/}} and report performance in \autoref{tab:pos-results}. 
The most frequently gapped PoS tags in our datasets correspond to closed word classes (such as ADP, DET, SCONJ, AUX, etc.), which is expected given that our open cloze tests are mostly focused on testing grammar rather than vocabulary. The best predicted classes, however, are NUM, SCONJ, NOUN, ADJ and INTJ which on closer inspection turn out to be very restricted classes: NUM includes only the word \textit{one}, INTJ only the word \textit{like}, SCONJ only a few subordinating conjunctions while NOUN and ADJ, despite being open classes, are limited to words used in common constructions such as \textit{order} (\textit{in order to}) or \textit{same} (\textit{the same}).

The two worst performing classes are PART (the particles \textit{to} and \textit{not}) and AUX (auxiliary verbs) and, once again, we conjecture that these words are so common in the language and in non-gapped positions that the model is unable to get them right most of the time. 
The remaining PoS classes vary in performance but we found only very weak correlation between PoS gap frequency in the test set and F\textsubscript{1} scores (Pearson's $r=0.1932$, Spearman's $\rho=0.1350$).

When we look at human annotations on the test set, however, performance by PoS is consistently higher and more even across the board. If we require that gaps are rated `good' by at least two annotators, accuracy values range between $75\%$ and $100\%$ for all PoS, with a mean of $85\%$.

Under these conditions, the best performing classes are NOUN ($100\%$), INTJ ($100\%$) and ADJ ($95\%$), which agree with automatic evaluation. Out of these, only NOUN achieves perfect accuracy across all annotators. The worst performing classes are PRON ($77\%$), NUM ($77\%$) and VERB ($75\%$) as opposed to the previous AUX and PART counterparts (now $79\%$ and $83\%$ respectively). When we require agreement by all annotators, the worst overall class is CCONJ with $44\%$.

\subsection{Qualitative Analysis}

\autoref{fig:heatmap} shows the output of our model for a sample text passage, where darker red indicates higher confidence in inserting a gap. The final model's predictions have a black frame (\textit{at}, \textit{in}, \textit{so}, \textit{after}, etc.) while the gold standard gaps are in yellow font (\textit{at}, \textit{in}, \textit{so}, etc.). There are $8$ matched gaps out of $11$ in this example, yielding $72.73\%$ accuracy. 

As can be seen in the figure, our model is able to identify appropriate gap candidates, even if they do not match the gold standard. In fact, annotators considered all the unmatched gaps in this example (\textit{after}, \textit{for} and \textit{take}) to be good and the second matched gap (\textit{in}) to be inappropriate. 
It is also interesting to see how the model prioritises function words and content words that are highly restricted in context (such as \textit{take} or \textit{part}), skilfully avoiding general gaps that could accept multiple answers and would be less effective for testing purposes.

\section{Conclusion and Future Work}

We described the first transformer-based approach to open cloze test generation. Our ELECTRA-based model is trained on two objectives: token classification (gap/non-gap) and language modelling (for predicting the expected answer). The model is further improved by manipulating the loss function and post-processing the results.

System accuracy using automatic evaluation is $53.89\%$ while human evaluation ranges between $75\%-82\%$, showing that at least $7$ out of $10$ gaps predicted are considered useful by experts. A detailed analysis of results reveals a few structural problems such as gaps in close proximity and inappropriate difficulty, which we plan to address in future work. Our test data and human annotations are released with this paper.


\if{false}
\begin{table}
\centering
\begin{tabular}{llllll}
\hline
 & P & R & F1 \\
\hline
Best model & 45.6 &	63.33	& 53.02 
\\
+ post-processing & - & - &  53.89 \\
\hline
\end{tabular}
\caption{\label{tab:test-results}
Results of the best model (multi-objective ELECTRA with loss manipulation) on the CEP blind test data. 
}
\end{table}
\fi

\if{false}
\begin{table}
\centering
\begin{tabular}{llllllllc}
\hline
 & P & R & F1 & PP-F1 \\
\hline
Best model & 0.456 &	0.6333	& 0.5302 &	
  53.89 \\
\hline
\end{tabular}
\caption{\label{tab:test-results}
Results of the best model (multi-objective ELECTRA with loss manipulation) on the CEP blind test data. PP: post-processed; PP-F1 is the performance after applying post-processing and fixing the number of predicted gaps to the number of gaps in the gold standard.
}
\end{table}
\fi

\if{false}
\begin{table}[t!]
\centering
\small
\setlength{\tabcolsep}{2.5pt}
	\begin{tabular}{|l|r|rrr|rrr|}
		\hline
        & \textbf{Test} &       \multicolumn{3}{c|}{\textbf{Micro-averaged}} & \multicolumn{3}{c|}{\textbf{Macro-averaged}} \\
        \cline{3-8}
		& \textbf{cases} & \textbf{P} & \textbf{R} & \textbf{F\textsubscript{1}} & \textbf{P} & \textbf{R} & \textbf{F\textsubscript{1}} \\ 
		\hline
        Unseen & 4 & 0.00 & 0.00 & 0.00 & 0.00 & 0.00 & 0.00 \\
        Low & 4 & 100.00 & 25.00 & 40.00 & 100.00 & 25.00 & 40.00 \\
        Medium & 30 & 36.11 & 43.33 & 39.39 & 27.38 & 38.71 & 32.00 \\
        High & 322 & 55.90 & 55.90 & 55.90 & 50.95 & 52.37 & 48.84 \\
        \hline
	\end{tabular}
	\caption{\label{tab:freq-results} Performance by gapped word frequency based on automatic evaluation.}
	\vspace*{-\baselineskip}
\end{table}
\fi

\section*{Acknowledgements}

The authors are immensely grateful to Louise Gilbert, Sally Moore and Clare Williams from  CUP\&A for their annotations.
This paper reports on research supported by Cambridge University Press \& Assessment, University of Cambridge.

\bibliography{references}
\bibliographystyle{acl_natbib}


\newpage
\clearpage

\appendix
\counterwithin{figure}{section}
\counterwithin{table}{section}
\renewcommand\textfraction{.1}

\onecolumn
\begin{multicols}{2}

\section{Dataset composition}
\label{app:dataset}

\begin{table}[H]
    \centering
    \begin{tabular}{c|rrr}
        \textbf{\# Gaps} & \textbf{Train} & \textbf{Dev} & \textbf{Test}  \\ \hline
        8 & 4 & 0 & 0 \\
        9 & 39 & 2 & 9 \\
        10 & 71 & 2 & 18 \\
        11 & 38 & 12 & 9 \\
        12 & 4 & 6 & 0 \\
        13 & 58 & 9 & 0 \\
        14 & 10 & 2 & 0 \\
        15 & 0 & 0 & 0 \\
        16 & 132 & 25 & 0 \\
        \hline
        \textbf{Total} & 356 & 58 & 36 \\
    \end{tabular}
    \caption{\label{tab:gaps} Distribution of the number of gaps per task in each section of the data.}
\vspace{-0.5\baselineskip}
\end{table}

\begin{table}[H]
    \centering
    \begin{tabular}{c|rrr}
        \textbf{\# Answers} & \textbf{Train} & \textbf{Dev} & \textbf{Test}  \\ \hline
        1 & 3637 & 639 & 296 \\
        2 & 689 & 111 & 45 \\
        3 & 147 & 23 & 16 \\
        4 & 77 & 11 & 2 \\
        5 & 11 & 3 & 1 \\
        6 & 3 & 0 & 0 \\
        7 & 1 & 0 & 0 \\
        \hline
        \textbf{Total} & 4565 & 787 & 360
    \end{tabular}
    \caption{\label{tab:answers} Distribution of the number of answers per gap in each section of the data.}
\vspace{-0.5\baselineskip}
\end{table}

\vfill\null
\columnbreak

\section{Model parameters}
\label{app:paramaters}

\begin{table}[H]
    \centering
    \begin{adjustbox}{width=1\columnwidth}
    \begin{tabular}{|l|c|c|}
    \hline
    Parameters & BERT & \makecell{Multi-objective\\ELECTRA} \\
    \hline
    Learning rate & $3\times10^{-5}$ & $3\times10^{-5}$ \\
    \hline
    Batch size & 1 & 1 \\
    \hline
    Number of epochs & 4 & 4 \\
    \hline
    Training steps & ${\dfrac{n}{b}}\times{e}$ & ${\dfrac{n}{b}}\times{e}$ \\
    \hline
    \end{tabular}
    \end{adjustbox}
    \caption{\label{tab:parameters} Model parameters used for the experiments. $n$: the number of training examples; b: batch size; e: number of epochs.}
\end{table}
\end{multicols}

\vspace{-5ex}
\section{Human labelling examples}
\label{app:reasons}

\begin{table*}[bht]
    \centering
    \small
    \begin{adjustbox}{width=1\columnwidth}
    \begin{tabular}{lp{5cm}p{5cm}}
        \textbf{Reason for rejection} & 
        \multicolumn{1}{c}{\textbf{Example gap in context}} & 
        \multicolumn{1}{c}{\textbf{Annotator comments}} \\ \hline
        Too close to other gaps &  ... the thousands of questions I asked \uline{as} a child \uline{were} met not by impatient answers ... & Minimum distance is not met. \\
        \hline
        Too many possible answers & ... and \uline{does} not sound threatening. & Many verbs could fit in this gap: \textit{does}, \textit{may}, \textit{might}, \textit{should}, \textit{will}, etc. \\
        \hline
        Too many gaps of this type & ... the country \uline{where} the largest number of bamboo varieties grow naturally... & Too many relative pronouns are tested in the task. \\
        \hline
        Answers can change meaning & ... the petals were narrower and \uline{less} clearly separated ... & The word \textit{more} also fits.\\
        \hline
        Answers can have different PoS & The Indian bansuri bamboo flute, \uline{when} played by a master musician, ...  & Other possible answers are \textit{often}, \textit{usually}, \textit{normally}, etc. \\ 
        \hline
        Gap depends on another & \uline{What} I love most about being on a horse \uline{is} that ... & The second gap depends on the first.\\
        \hline
        Repeated gap &  ..., \uline{although} she later became a biologist. & The task has another gap where \textit{although} is a possible answer.\\
        \hline
        Phantom gap & The name actually refers to the statuette \uline{which} all of the winners receive. & \textit{Which} can be omitted.\\
        \hline
        Unacceptable outlier & The school was by \uline{no} means an overnight success; ... & The phrase \textit{by no means} is at the C1 CEFR level.\\
        \hline
        Other (please specify) & \uline{It} is sometimes said that animals use language. & Avoid having a gap for the very first word in the text. \\
        \hline
    \end{tabular}
    \end{adjustbox}
    \caption{\label{tab:reasons} Example of the different reasons given by the annotators for rejecting a gap proposed by our system.}
\end{table*}

\end{document}